\documentclass[letterpaper, 10 pt, conference]{ieeeconf}  % Comment this line out if you need a4paper

\IEEEoverridecommandlockouts                              % This command is only needed if 
\usepackage[noadjust]{cite}

\usepackage[utf8]{inputenc} % allow utf-8 input
\usepackage[T1]{fontenc}    % use 8-bit T1 fonts
\usepackage{hyperref}       % hyperlinks
\usepackage{url}            % simple URL typesetting
\usepackage{booktabs}       % professional-quality tables
\usepackage{amsfonts}       % blackboard math symbols
\usepackage{nicefrac}       % compact symbols for 1/2, etc.
\usepackage{microtype}      % microtypography
\usepackage{xcolor}         % colors
\usepackage{graphics}
\usepackage{graphicx}
\let\labelindent\relax % Needed for IEEE template
\usepackage{enumitem}
\usepackage{caption}
\usepackage{xspace}
\usepackage{siunitx}
\usepackage{caption}
\usepackage{subcaption}
\usepackage{wrapfig}
\usepackage{multirow}
\usepackage{tabularx}
\usepackage{float}

\usepackage{wrapfig}

\definecolor{MyDarkBlue}{rgb}{0,0.08,1}
\definecolor{MyDarkGreen}{rgb}{0.02,0.6,0.02}
\definecolor{MyDarkRed}{rgb}{0.8,0.02,0.02}
\definecolor{MyDarkOrange}{rgb}{0.40,0.2,0.02}
\definecolor{MyPurple}{RGB}{111,0,255}
\definecolor{MyRed}{rgb}{1.0,0.0,0.0}
\definecolor{MyGold}{rgb}{0.75,0.6,0.12}
\definecolor{MyDarkgray}{rgb}{0.66, 0.66, 0.66}

\newcommand{\baselineone}{Kipf et al.~\cite{kipf2017gcn}\xspace}
\newcommand{\baselinetwo}{Yang et al.~\cite{yang2018visual}\xspace}

\newcommand{\experimentone}{relational object choice\xspace}
\newcommand{\Experimentone}{Relational object choice\xspace}

\newcommand{\experimenttwo}{directed object navigation\xspace}
\newcommand{\Experimenttwo}{Directed object navigation\xspace}

\newcommand{\experimentthree}{exploratory object navigation\xspace}
\newcommand{\Experimentthree}{Exploratory object navigation\xspace}

\newcommand{\experimenttwothree}{directed and exploratory object navigation\xspace}

\newcommand{\sgnoattn}{RGB-D + SG\xspace}
\newcommand{\sgattn}{RGB-D + SG + TD ATTN\xspace}
\newcommand{\sgnoattnpm}{PM: RGB-D + SG\xspace}
\newcommand{\sgattnpm}{PM: RGB-D + SG + TD ATTN\xspace}

\title{\LARGE \bf
Task-Driven Graph Attention for \\ Hierarchical Relational Object Navigation}

\author{Michael Lingelbach$^{1}$, Chengshu Li$^{1}$, Minjune Hwang$^{1}$, Andrey Kurenkov$^{1}$, Alan Lou$^{1}$, \\ Roberto Martín-Martín$^{2}$, Ruohan Zhang$^{1}$, Li Fei-Fei$^{1}$, Jiajun Wu$^{1}$% <-this % stops a space
\thanks{Department of Computer Science, $^{1}$~Stanford University, CA, USA, $^{2}$~University of Texas at Austin, TX, USA,
        {\tt\small mjlbach@stanford.edu}}%
}

\begin{document}

\maketitle
\thispagestyle{empty}
\pagestyle{empty}

%%%%%%%%%%%%%%%%%%%%%%%%%%%%%%%%%%%%%%%%%%%%%%%%%%%%%%%%%%%%%%%%%%%%%%%%%%%%%%%%

\begin{abstract}
Embodied AI agents in large scenes often need to navigate to find objects. In this work, we study a naturally emerging variant of the object navigation task, hierarchical relational object navigation (HRON), where the goal is to find objects specified by logical predicates organized in a hierarchical structure---objects related to furniture and then to rooms---such as finding an apple on top of a table in the kitchen. Solving such a task requires an efficient representation to reason about object relations and correlate the relations in the environment and in the task goal. HRON in large scenes (e.g. homes) is particularly challenging due to its partial observability and long horizon, which invites solutions that can compactly store the past information while effectively exploring the scene. We demonstrate experimentally that scene graphs are the best-suited representation compared to conventional representations such as images or 2D maps. We propose a solution that uses scene graphs as part of its input and integrates graph neural networks as its backbone, with an integrated task-driven attention mechanism, and demonstrate its better scalability and learning efficiency than state-of-the-art baselines.\end{abstract}

% \jw{No, not scene graphs, but task-oriented scene graphs. It's a new representation. Different from scene graphs.}

\section{Introduction}
Searching for objects in large scenes is a challenging component of many embodied AI activities such as rearrangement tasks~\cite{batra2020rearrangement,gan2021threedworld,weihs2021visual} and household activities~\cite{srivastava2022behavior,wisspeintner2009robocup}. When searching for an object, the embodied AI agent needs to navigate across different rooms and uses what it observes along the way to make optimal navigation decisions until it finds the target. Given the complexity of natural scenes with multiple rooms and objects, a challenge in object navigation is to devise a scalable solution that efficiently represents and exploits known information for future decisions.

Most prior object search work focused on the version of the problem defined as finding \textit{any} instance of the target object category (e.g., ``find \textit{any} pair of shoes'') -- namely the \textbf{object navigation} problem \cite{zhu2017target,anderson2018evaluation,batra2020objectnav}. An alternative definition requires finding a specific \textit{instance} of the target object category, e.g., ``the old shoes''.
This task has been called \textbf{instance object navigation}~\cite{li2021ion} or ION. ION is oftentimes a more natural problem definition, as realistic downstream tasks usually require a specific object instance (``the red book'') rather than any instance (``any book''). In this work, we focus on a new instantiation of ION that introduces additional \textit{hierarchical relational} constraints in the definition (object-furniture, furniture-room), such as ``find the shoes under the bed in the bedroom'', or ``find the mug on the table in the kitchen''. We call this problem \textbf{hierarchical relational object navigation} or \textbf{HRON}.

% representation, solution
An ideal solution to an object navigation task should combine an appropriate input representation with an optimal mechanism to extract the necessary information to guide navigation; these can be task-dependent.
While egocentric RGB-D images, segmented images, point clouds, and their integration into 2D or 3D (semantic) maps have been successfully used as inputs to object navigation~\cite{chaplot2020object,wahid2020learning,ye2021auxiliary}, solutions that use them as input perform poorly for HRON (see Sec.~\ref{s:results}). This is because they cannot effectively represent relational information nor scale to large, multi-room natural scenes.
In contrast, scene graphs~\cite{johnson2015image,armeni20193d,ravichandran2021hierarchical,hughes2022hydra} -- graphs where nodes are objects or rooms, and edges are pairwise relations between them -- provide a compact scene representation that captures the critical information to guide a HRON solution. 
Therefore, our proposed HRON solution uses a scene graph built during exploration as the input representation. 

% GNN
The information encoded in graph structures can be extracted and leveraged efficiently using graph neural networks (GNNs)~\cite{scarselli2008graph, kipf2017gcn,wu2020comprehensive,ravichandran2021hierarchical, Savarese-RSS-19, yang2018visual, zhu2021soon}. However, GNNs' success may be limited if they are applied naively to large graphs with irrelevant nodes and edges like the ones representing realistic scenes with hundreds of objects.
% For instance, a GNN may not be beneficial for relational object navigation when most scene objects are not relevant to the current goal -- which is often the case.
Hence, we propose to integrate task-conditioned attention into GNNs to focus on the task-relevant elements of the graph in order to better aggregate their features and solve the current task. 

% contributions
In summary, our contributions are threefold:
\begin{itemize}[leftmargin=5.5mm]
    \item We introduce the hierarchical relational object navigation (HRON) task. HRON requires more sophisticated reasoning about object and room relations than object navigation and instance object navigation.%, and approaches that scale to larger scenes. 
    \item We propose a novel solution to HRON based on a scene graph representation, that combines graph neural networks and task-driven attention for better scalability and learning efficiency for HRON in large scenes. Through experimental evaluation, we show that a reinforcement learning (RL) agent with our proposed architecture outperforms prior work with better performance and sample efficiency.
    \item We introduce concrete instantiations of HRON
    in three tasks of increasing complexity and realism. We provide a symbolic implementation for the first task and a physically-grounded implementation for the other two tasks in iGibson 2.0 ~\cite{li2021igibson} -- a 3D simulator that provides photorealistic rendering and physics simulations in large household scenes. These environments will be publicly accessible for future research.
\end{itemize}

\section{Related Work}

\begin{figure*}[ht]
\vspace{0.25cm}
\centering
    \includegraphics[width=0.85\textwidth]{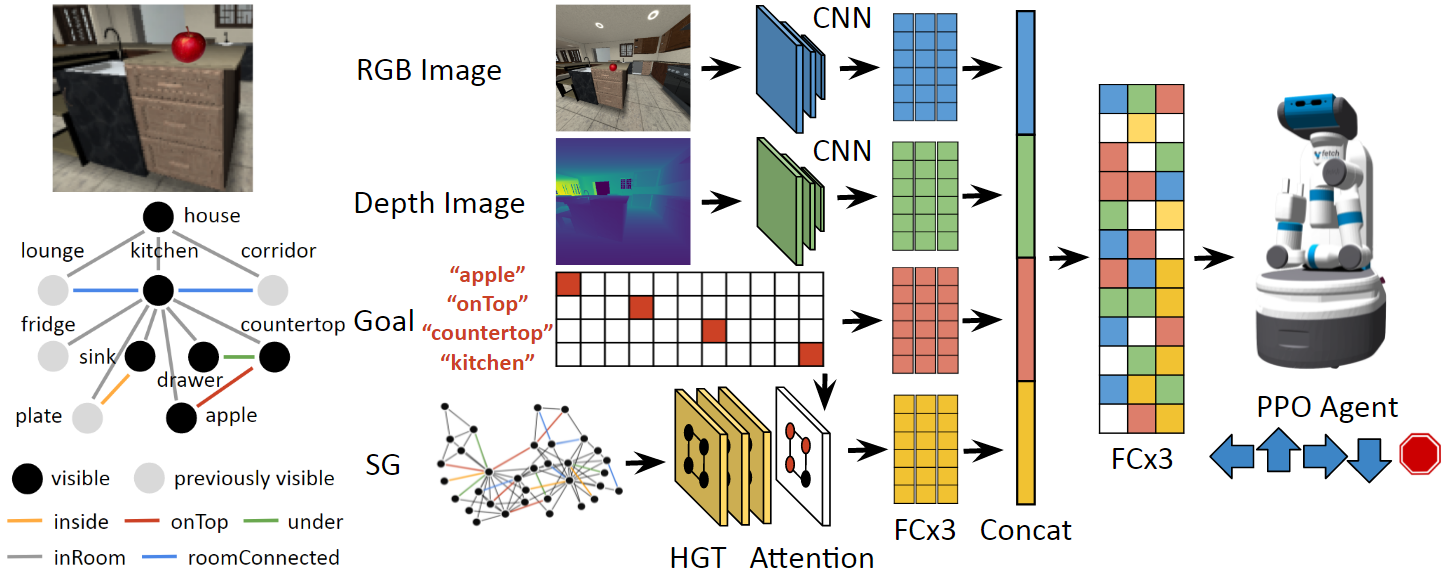}
    \caption{Overview of our proposed model. We incrementally build a Scene Graph (SG) from RGB-D images where each node represents an object or a room, and each edge represents a physical relation between them. The SG is processed by a Heterogeneous Graph Transformer (HGT) with a task-driven attention mechanism to extract task-relevant information. The model also receives the current RGB-D image and a one-hot encoded goal description, which are processed by their respective learned encoders. All the features are then concatenated together before being fed to more fully connected (FC) layers, and eventually to a PPO agent that outputs navigation actions.}
    \label{fig:architecture}
    \vspace{-0.5cm}
\end{figure*}

% visual object search, object nav, etc
\paragraph{Object navigation}
% problem formulation differences

Embodied search for objects with partial observability is a long-studied problem in embodied AI and robotics~\cite{Tejwani1989RobotV,wixson1994using,baezayates1993searching, forssen2008informed,ekvall2007object,fekete2006online,sjo2009object,yang2018visual,mishkin2019benchmarking,mousavian2019visual,wortsman2019learning,chaplot2020object,du2020learning,li2020unsupervised,staroverov2020real,wani2020multion,du2020vtnet,zhang2021hierarchical,pal2021learning,zhang2021hierarchical,zhu2021soon,zhu2021deep}. Over the past several years many works have focused on learning-based approaches, since such approaches require less prior knowledge in novel environments \cite{ye2020seeing}. However, they are typically limited to searching in single rooms only containing one instance of the target object~\cite{batra2020objectnav}. %Therefore, they do not lend themselves well to large, natural environments in which more than one instance of a given object type exists.
Recently, ION (Instance-level Object Navigation)~\cite{li2021ion} and SOON (Scenario Oriented Object Navigation)~\cite{zhu2021soon} expanded the problem definition to locating an instance of an object with specific attributes and relations. However, ION still focuses on small single-room scenes, and SOON focuses on discrete selection over pre-defined waypoints trained with privileged information rather than embodied navigation. In contrast, our focus is on long-horizon learned embodied navigation in a large, multi-room (hierarchical) scene. We therefore choose to focus on defining object instances strictly in terms of relational constraints, since these constraints are naturally suited to directing the agent towards certain areas in larger scenes. A comparison between different problem setups can be seen in Table~\ref{tab:problems}. 

\begin{table*}
    \vspace{0.2cm}
    \centering
    \resizebox{0.93\textwidth}{!}{
    \begin{tabular}{c|c c c c}
         \toprule
         &  ON~\cite{batra2020objectnav} & ION~\cite{li2021ion} & SOON~\cite{zhu2021soon} & HRON (Ours)\\
         \midrule
         \multirow{4}{*}{} 
          Goal & cup &  blue, plastic & cylindrical, metallic and tall lamp which is set in the & apple  \\
          (Example) &  &  cup  &  bright living room...The living room is on the first  & on top of counter\\
          &  & near toaster & floor, next to the dining room and next to the kitchen  & in the kitchen  \\ 
         \midrule
         \multirow{2}{*}{Environment}  & Habitat \cite{savva2019habitat} & AI2-Thor \cite{kolve2017ai2} & Matterport3D \cite{chang2017matterport3d} & iGibson 2.0 \cite{li2021igibson} \\
          & single-room & single-room & multi-room & multi-room\\
         \midrule
         Embodiment & navigation commands & navigation commands & node choice in a pre-defined graph & navigation commands  \\
         Task horizon & vary & $\sim$45 steps & $\sim$10 steps & $ \sim$100 steps \\
         \bottomrule
    \end{tabular}}
    \caption{A comparison between different instances of object navigation (ON) tasks. HRON focuses on long-horizon learned embodied navigation in a large, multi-room (hierarchical) scene.}
    \label{tab:problems}
    \vspace{-0.5cm}
\end{table*}

% scene graphs
\paragraph{Scene graphs in object navigation}
Although learning a navigation policy directly from sensor input is viable~\cite{ye2021auxiliary}, additional inductive biases and knowledge representations improve efficiency in larger scenes. Unlike other representations such as semantic maps, scene graphs scale with the number of objects rather than the size of the scene, making them suitable as a knowledge and memory representation for object navigation in large-scale scenes~\cite{amiri2022reasoning, kumar2021gcexp, santos2022deep}. 
Scene graphs explicitly and compactly store information about objects' geometry, placement, semantics, and relationships, which makes them ideal for reasoning about object relations. Furthermore, graphs naturally encode hierarchical relations and have been utilized in several prior works~\cite{ravichandran2021hierarchical,pal2021learning,zhu2021hierarchical,kurenkov2021semantic}. However, while prior works focused on using scene graphs for the standard object navigation task with homogeneous graph edges, our focus is on studying how to best leverage scene graphs for HRON problems with directed heterogeneous edge types. Relational information in the edges provides useful information for targeted exploration of the scene. For example, when searching for an apple on a table in the kitchen, the presence of an "in room" edge between a particular table instance and a kitchen instance indicates this table should be prioritized during exploration. 

\paragraph{Attention in object navigation}
The importance of attention in human visual search has been long recognized~\cite{sperling1978attention,bravo1992role,wolfe2017five}. Developing an attention mechanism for embodied AI agents is an active research topic. Such attention can be in the form of saliency map on egocentric RGB images~\cite{mayo2021visual}, or weights on 2D maps~\cite{seymour2021maast, ramakrishnan2021exploration}. Attention can potentially help leverage large scene graphs in visual search, but an appropriate attention mechanism still needs to be explored since there are many forms of attention in graph neural networks developed for different purposes~\cite{velivckovic2017graph,yun2019graph,kim2022find}. Our work builds upon existing research and proposes a framework for incorporating task-driven attention into a scene graph representation for the challenging HRON problem.

\section{Method}
\label{sec:method}

Our solution for HRON comprises four elements (Fig.~\ref{fig:architecture}). First, at the core is a scene graph representation of the environment, incrementally constructed from current RGB-D images during task execution, and used as part of the input to our model (Sec.~\ref{sec:sg}). Second, a graph neural network and task-driven attention module are used to summarize the information of the scene graph into a single graph feature (Sec.~\ref{sec:gnn}). Third, visual RGB-D inputs and the goal description, a tuple of one-hot encodings, are converted to feature vectors through learned layers, and all the input vectors are fused into a single vector (Sec.~\ref{sec:fusion}). Finally, this fused vector is the learned representation used by a reinforcement learning agent that trains through interaction with the environment to find objects specified by hierarchical relational constraints (Sec.~\ref{sec:policy}). We then describe each component in detail.

\subsection{Scene graph representation}
\label{sec:sg}

A scene graph is comprised of a set of nodes and a set of directed edges, where each node represents a physical entity (e.g. objects, rooms). The node feature includes the attributes and the states of the physical entity (e.g. semantic class, 3D pose, size). All the nodes have their poses defined in the local coordinate frame of the agent, such that the agent knows its own position in the scene graph. The directed edges represent physical relations between the entities, e.g. ``roomConnected'' (room-room relation),  ``onTop'' (object-object relation), and ``inRoom'' (object-room relation).

To build the scene graph, our method extracts information from the RGB-D images using a perfect object detector provided by the iGibson simulator at each step of the navigation task and accumulates it in an incremental fashion. In simulation, we emulate the execution of a scene graph building method such as the ones presented in prior work~\cite{rosinol20203d,armeni20193d,rosinol20203d}. 
At any given step of the search, objects and rooms that are within the agent's field of view but not yet in the scene graph are added to it as new nodes, and those that already exist will have their node features updated. The edges that connect to these nodes will be detected and updated as well. 
This procedure replicates a realistic incremental graph-building process in an embodied AI navigation agent.

\subsection{Graph neural network architecture}
\label{sec:gnn}

\paragraph{Heterogeneous graph transformer (HGT)} To compute a per node embedding, the graph is passed through three heterogeneous graph transformer layers~\cite{DBLP:journals/corr/abs-2003-01332} with ReLU activations. The HGT convolutions use distinct edge-based matrices for each edge type when computing attention, allowing the model to learn representations conditioned on different edge types, rather than connectivity alone. 

\paragraph{Graph attention pooling} The final pooling layer applies a weighted mean pooling over all the node embeddings with a task-driven attention mechanism to create a single vector summarizing the scene graph. In order to effectively aggregate task-relevant nodes, the task-driven attention mechanism assigns a weight of 1 to all nodes for which the semantic category matches any semantic category found in the current episodic goal description, and 0 otherwise. .

\subsection{Multimodal feature fusion} To fuse features of global/history info (scene graphs), local info (current RGB-D images), and goal info, we process the RGB-D images with three consecutive convolutional layers, flattening, and three fully-connected (FC) layers, and process the one-hot encoded goal description with three FC layers. We then concatenate the embeddings from all three branches (including the aforementioned scene graph branch) and pass the concatenated feature through an additional stack of three FC layers for feature fusion.

\label{sec:fusion}

\subsection{Policy training} 
\label{sec:policy}
The model is trained end-to-end using a PPO~\cite{schulman2017proximal} implementation adapted from RLlib~\cite{liang2018rllib}, across 8 parallel environments for approximately 1.5 million environment steps of experience. The fused feature vector mentioned above is passed through separate, dedicated 3-layer FC for policy and value networks, respectively. We found that it was important to have sufficient network depth for these two networks in order for the policy to leverage the graph features.

\section{Experimental Evaluation}

We design our experiments to answer these two questions:
\begin{itemize}[leftmargin=5.5mm]
    \item \textbf{Q1}: Does the scene graph representation help the agent learn faster and perform better in the HRON tasks?
    \item \textbf{Q2}: Does the attention mechanism facilitate learning in large, populated scenes?
\end{itemize}

\begin{figure*}
    \vspace{0.2cm}
    \centering
    \includegraphics[height=0.25\textwidth]{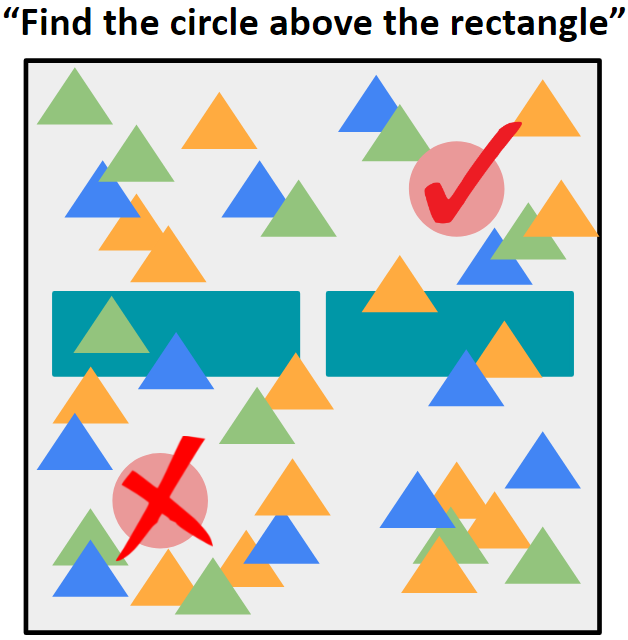}
    \includegraphics[height=0.25\textwidth]{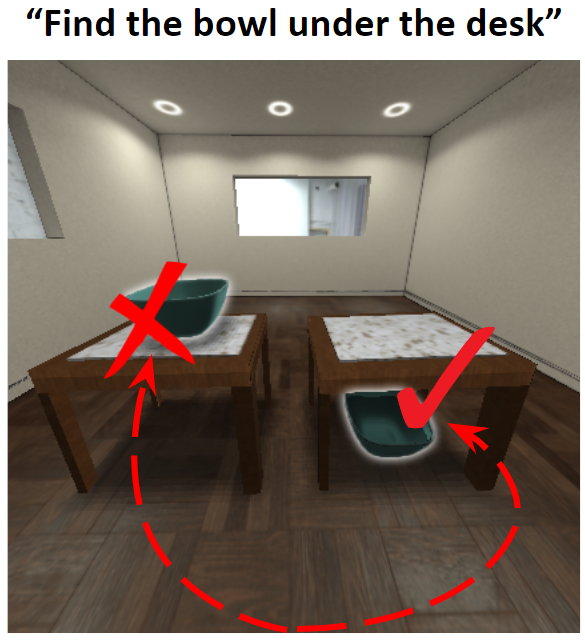}
    \includegraphics[height=0.25\textwidth]{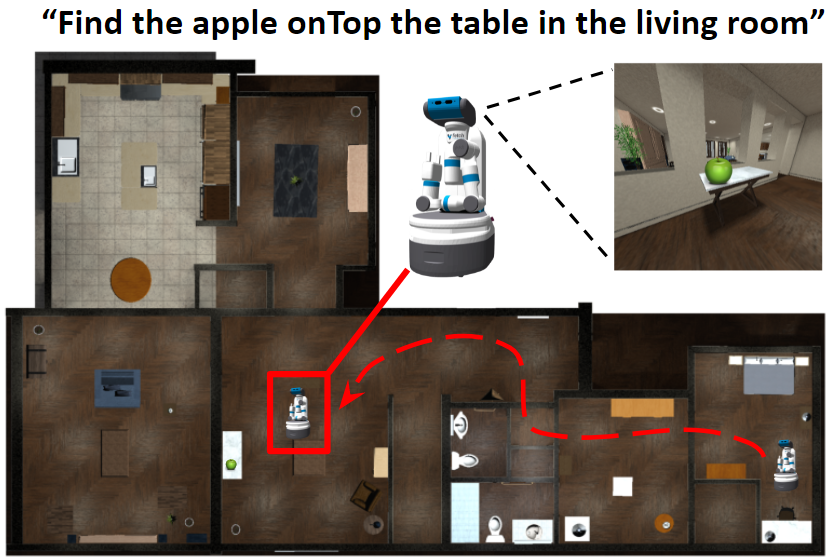}
    \caption{Three concrete tasks that illustrate relational object reasoning: \experimentone, \experimenttwo, and \experimentthree. In \experimentone~(left), given a relational object goal such as ``the circle above the rectangle'', the agent should output a binary choice (\texttt{left}/\texttt{right}) that corresponds to the side of the environment that satisfies the goal. In \experimenttwo and \experimentthree~(middle, right), given a relational object goal such as ``the apple on top of the table in the living room'', the agent should output a sequence of discrete navigation actions (\texttt{forward}/\texttt{backwards}/\texttt{left}/\texttt{right}/\texttt{stop}) to find and get close enough to the target object.}
    \label{fig:problem_setup}
    \vspace{-0.5cm}
\end{figure*}

\subsection{Experimental setup}
\label{sec:experimental_setup}
We design three tasks that illustrate different aspects of relational object reasoning with increasing complexity and realism (see Fig.~\ref{fig:problem_setup}):

\paragraph{\Experimentone} The agent is presented with a 2D environment with two circles and rectangles: one circle above and the other below their respective rectangle. The agent is given a goal description of ``circle above rectangle'' or ``circle below rectangle'' and must choose one of two possible actions (\texttt{left}/\texttt{right}) to select the side of the environment that satisfies the goal description. The agent is given a reward of 1 if it selects the correct action, and a reward of 0 otherwise. The episode terminates after one step (bandit problem). To study the effect of high scene complexity, we add a random number (up to 75) of triangles as distractors. 

The other two tasks (\experimenttwothree) are implemented in iGibson 2.0~\cite{li2021igibson}, which provides photorealistic rendering and accurate physics simulation.
with an onboard RGB-D camera. All assets (scenes and objects) from iGibson 2.0~\cite{li2021igibson} can be freely used within the iGibson simulator. Both the assets and the simulator are publicly available from \href{http://svl.stanford.edu/igibson/}{their website} under MIT license.

\paragraph{\Experimenttwo} The agent is presented with a symmetrically arranged room, where an object (``bowl'', ``gym shoe'', ``apple'') is spawned according to a relational state (``onTop'', ``inside'', ``under'') with respect to an associated piece of furniture (``shelf'', ``table'') on each half of the room, differing only in terms of relational state. The agent is initialized at the same starting position on each episode. The observation space includes 1) RGB-D images from onboard sensors, and 2) tokenized, one-hot encoded goal descriptions (object-relation-furniture). The agent outputs one of the five discrete navigation actions: \texttt{forward} (\SI{0.2}{\meter}), \texttt{backward} (\SI{0.2}{\meter}), \texttt{left} (turn 30 degrees), \texttt{right} (30 degrees), and \texttt{stop}, which will be physically executed. The agent achieves success if it navigates within a fixed distance of the goal object ($d=1$m). The episode terminates if the agent runs out of time, with a maximum episode length of 500 timesteps, or equivalently, 50 simulated seconds, or if the agent approaches the incorrect object ($d=\SI{1}{\meter}$). The agent is given a reward of $10$ if it achieves success, a reward of $-5$ if it approaches the incorrect object, and a reward of $0$ otherwise. The agent is also provided with a geodesic-distance-based reward that encourages the agent to approach the goal object.

\paragraph{\Experimentthree} The agent is sampled randomly in one of the rooms in the realistic \texttt{Wainscott\_0\_int} iGibson scene~\cite{shen2020igibson} populated with furniture. The goal object category, relational state, associated furniture category, and room category of the target object are randomly selected at the beginning of each episode, e.g. ``apple on top of the table in the living room''. An instance of an object model matching the object category is sampled to fulfill the given relational constraint in a physically stable manner (e.g., an apple is placed on top of an instance model of a table in the living room). The observation space, action space, reward function, and termination conditions are identical to those of \experimenttwo\ with two exceptions: 1) the goal description contains hierarchical relational constraints: object-relation-furniture and furniture-inRoom-room, 2) the episode does not terminate, and the agent doesn't receive a negative reward when the agent approaches incorrect objects.

\subsection{Baselines and ablation studies} We compare our method against two state-of-the-art baselines, Graph Convolution Network (GCN) by \textbf{Kipf et al.}~\cite{kipf2017gcn} and the visual navigation method using scene priors by \textbf{Yang et al.}~\cite{yang2018visual}, both from the object navigation literature. Since the code for these works is not publicly available, we tried our best to replicate their approaches.%~\footnote{Our implementation will be made available at acceptance, together with our method for other researchers to use}. 

\begin{itemize}[leftmargin=5.5mm]
\item \textbf{\baselineone}: Graph Convolutional Networks (GCNs) are a simpler form of graph neural network compared to HGT, as it does not handle heterogeneous edges and involves no attention. Both~\cite{yang2018visual} and~\cite{seymourgraphmapper}
use GCNs to process their graphs. We implement this baseline by replacing our HGT with GCNs, and otherwise keeping all other aspects the same as our method.

\item \textbf{\baselinetwo}: We replicate the approach used in~\cite{yang2018visual}. Different from our method, this baseline represents the goal as a fastText~\cite{joulin2016bag} vector for the target object category, and creates node features from the fastText vectors for the nodes' category and the ResNet-50~\cite{he2016deep} softmax encoding of the current RGB image.
\end{itemize}

To showcase the effectiveness of each component of our model, we also perform extensive ablation studies.
\begin{itemize}[leftmargin=5.5mm]
\item \textbf{RGB-D only}: The model only receives RGB-D images and one-hot encoded goal description (no scene graph).
\item \textbf{RGB-D + MM}: Instead of incrementally building a scene graph, the model uses the accumulated RGB-D images to build a 2D, top-down, semantic metric map. Similar to SemExp~\cite{chaplot2020object}, the model projects a semantically-segmented point cloud (extracted from depth images with ground-truth semantic class information from the simulator) unto a 2D, top-down map as an image, where each pixel represents the physical space of \SI{0.23}{\meter} $\times$ \SI{0.23}{\meter}. If two points are projected onto the same 2D grid, the point with a higher z-value takes priority. The scene graph branch in the original model is replaced with a metric map branch that is identical to the RGB and Depth branches (conv layers, flattening, and MLP layers).
\item \textbf{\sgnoattn}: The attention mechanism is removed.

\item \textbf{\sgattn}: This is our main model as described in Sec.~\ref{sec:method}.

\end{itemize}

We also experiment with a set of ablations that provide additional information to the agent at the beginning of the episode. Specifically, we assume that the agent performs a pre-mapping procedure of the scene and stores all the furniture in a metric map or a scene graph. In other words, instead of incrementally building metric maps or scene graphs from scratch for each episode, the agent is provided with a pre-mapped scene representation that does not contain the target object as it is yet to be discovered. We argue that this is an alternative realistic setup that appears naturally when the agent pre-maps or searches consecutive for objects in the same scene. We call these variants of the models \textbf{PM: RGB-D + MM} and \textbf{PM: RGB-D + SG}. We run the full set of baselines and ablations for the \experimentthree\ task and a subset of them for the other two tasks when appropriate.

\section{Results}
\label{s:results}

To answer the \textbf{Q1} and \textbf{Q2} raised in the previous section, our main finding is that scene graph representation does help embodied AI agents to perform better in tasks that require relational object reasoning. Moreover, in large, populated scenes, where scene graph size grows to around 100 nodes, the task-driven attention mechanism is essential for task performance since it significantly helps the aggregation of task-relevant information across nodes. 

For quantitative results, we report the Success Rate (SR), whether or not the agent successfully approaches the target object within the time limit, and the Success weighted by Path Length (SPL), a ratio of the agent path length to the optimal path length conditioned on task success~\cite{anderson2018spl}, averaged over 20 episodes across three random seeds.

\paragraph{\Experimentone}
From Table~\ref{tab:exp1}, we observed that when there are no distractors, both variants of our models quickly learn to output the correct action that matches the goal description. However, when there are a large number of distractors, our model without attention \textbf{SG} has a significant drop in performance. Our model with task-driven attention \textbf{SG + TD ATTN} can salvage most of the performance loss and achieve near-perfect success.

\begin{table}[t]

%\captionsetup{skip=0pt,width=.48\textwidth,single}
\vspace{0.2cm}
\captionsetup{justification=centering,singlelinecheck=false}
\caption{\Experimentone: \label{tab:exp1}Success Rate}
\centering
\begin{tabular}{lll}
    \toprule
    Distractors & Model & SR \\
    \midrule
    \multirow{2}{*}{No} & SG & 0.997$\pm$0.003 \\
    & SG + TD ATTN & 0.991$\pm$0.009 \\
    \midrule
    \multirow{2}{*}{Yes} & SG & 0.659$\pm$0.002\\
    & SG + TD ATTN & \textbf{0.993$\pm$0.007}\\
    \bottomrule
\end{tabular}
\vspace{0.25cm}
%}
\caption{\label{tab:exp2}\Experimenttwo: Success Rate and \\ 
Success weighted by Path Length}

\resizebox{0.99\columnwidth}{!}{
\begin{tabular}{lllll}
    \toprule
    {Distractors} & {Model} & {SR}$\uparrow$ & {SPL}$\uparrow$ \\
    \midrule
    \multirow{3}{*}{No} & RGB-D only & 0.450$\pm$0.141 & 0.197 $\pm$ 0.039 \\
    & \sgnoattn & \textbf{0.967$\pm$0.024} & \textbf{0.949 $\pm$ 0.001} \\
    & \sgattn & \textbf{0.983$\pm$0.024} & \textbf{0.958 $\pm$ 0.019} \\
    \midrule
    \multirow{3}{*}{Yes} & RGB-D only & 0.417$\pm$0.103 & 0.173 $\pm$ 0.105 \\
    & \sgnoattn & \textbf{0.425$\pm$0.025} & \textbf{0.252 $\pm$ 0.046} \\
    & \sgattn & \textbf{0.900$\pm$0.041} & \textbf{0.864 $\pm$ 0.039} \\
    \bottomrule
\end{tabular}
}
\vspace{-0.2cm}
\end{table}

\paragraph{\Experimenttwo} %From Fig.~\ref{fig:exp1_exp2} and 
From Table~\ref{tab:exp2}, we observed that the \textbf{RGB-D only} ablation is not able to solve the task, and achieves only chance-level performance. Similar to the previous task, when there are no distractors, both variants of our scene graph model \textbf{\sgnoattn} and \textbf{\sgattn} are able to quickly learn to ground the goal description into the scene graph and choose a sequence of navigation actions that bring the robot close to the correct goal object. With distractors, on the other hand, task-driven attention is still critical for the effective aggregation of task-relevant information.  %It is able to solve the task with nearly a 100\% success rate within 500,000 environment steps.

% \begin{figure*}
% \centering
% \includegraphics[width=0.32\textwidth]{figs/3a_caption_new.pdf}
% \hfill
% \includegraphics[width=0.32\textwidth]{figs/3b_caption_new.pdf}
% \hfill
% \includegraphics[width=0.32\textwidth]{figs/relational_target_nav_new.pdf}
% \caption{Success Rate (SR) versus environment steps over training for the \experimentone\ and \experimenttwo\ tasks. The left two plots depict the \experimentone\ task with and without distractors, respectively. We observed that without attention, the performance of our model suffers significantly when there are many distractors. The rightmost plot depicts the \experimenttwo\ task. We observed that the \textbf{\sgnoattn} performs much better than \textbf{RGB-D only}, suggesting that scene graphs are effective representation for tasks that require relational reasoning.}
% \label{fig:exp1_exp2}
% \end{figure*}

\paragraph{\Experimentthree} 

\begin{figure*}[t]
    \centering
    \includegraphics[width=0.32\textwidth]{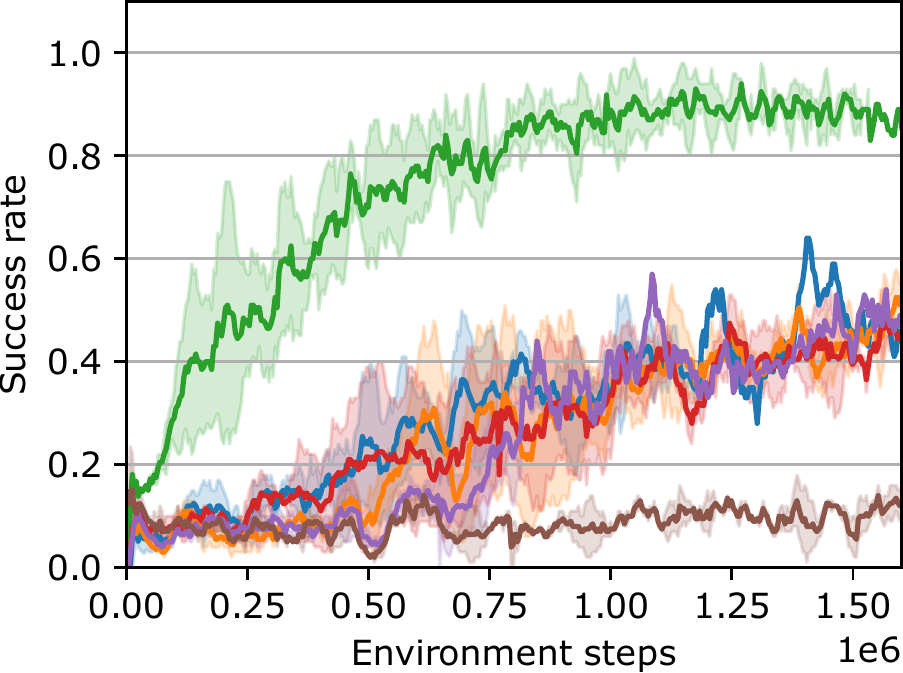}
    \includegraphics[width=0.32\textwidth]{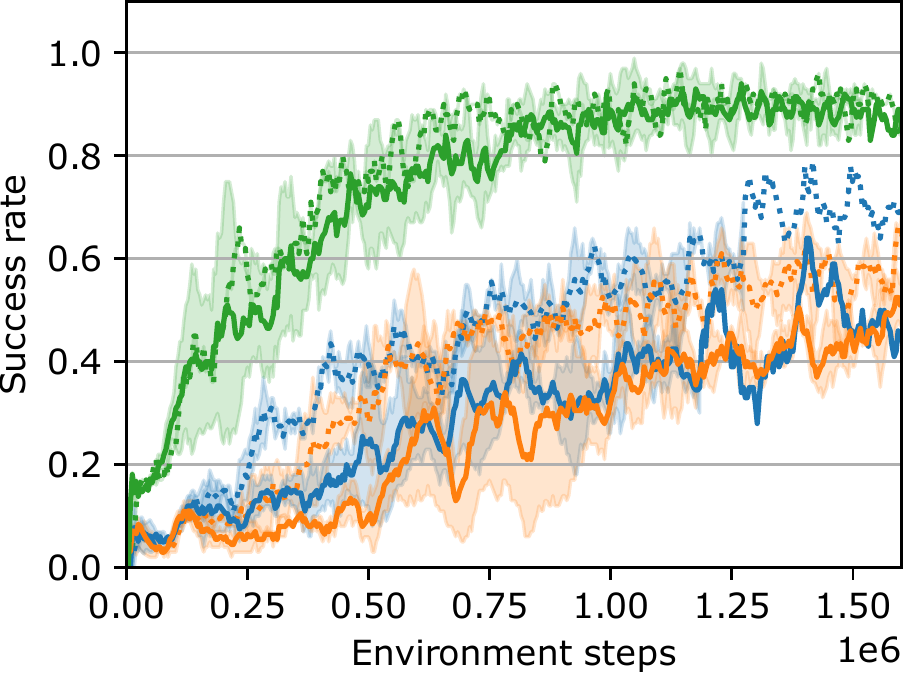}
    \includegraphics[width=0.16\textwidth, trim={13cm 10cm 10cm 0cm}, clip ]{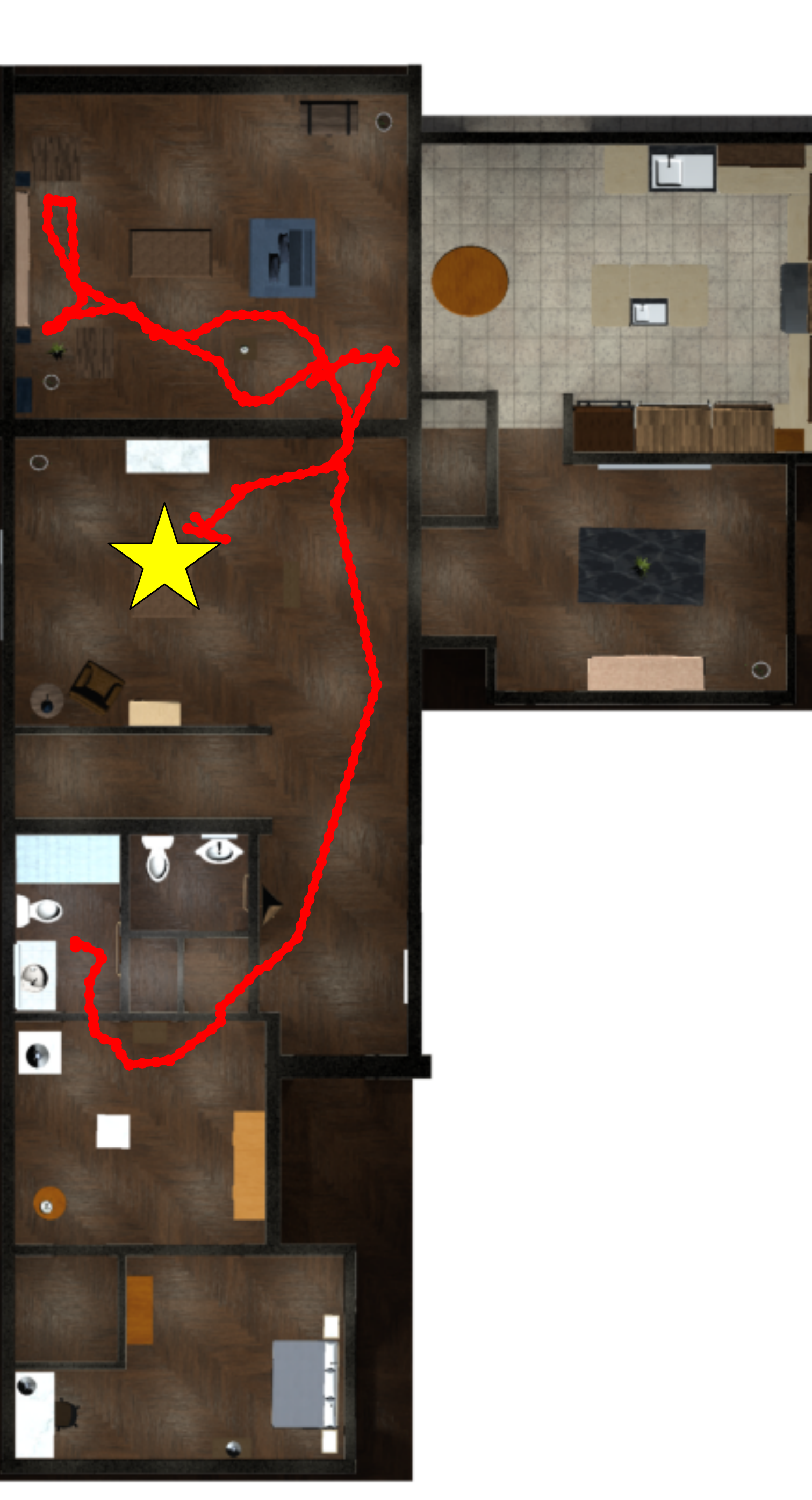}
    \includegraphics[width=0.16\textwidth, trim={13cm 10cm 10cm 0cm}, clip ]{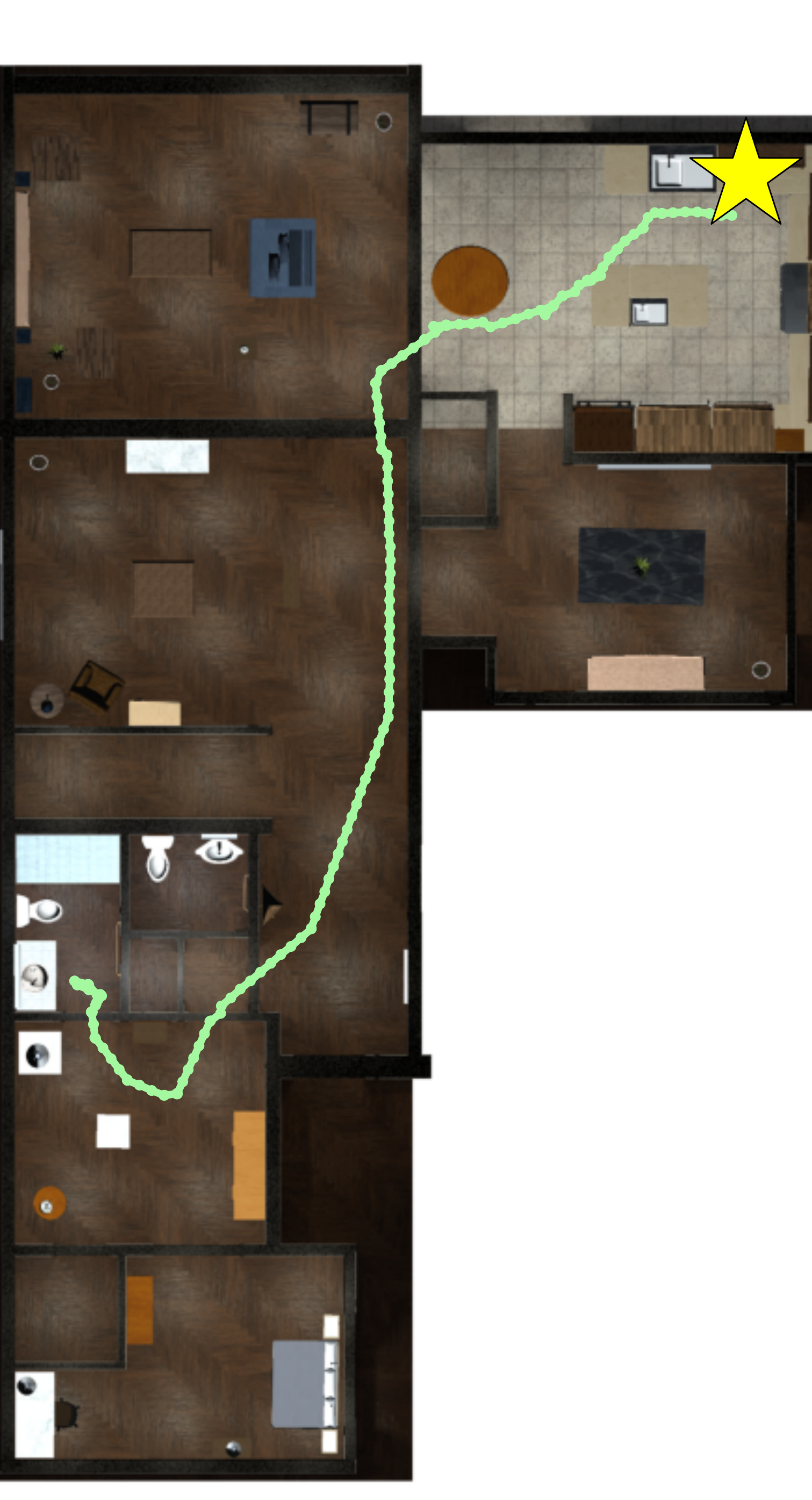}
    \raggedright
    \includegraphics[width=0.66\textwidth]{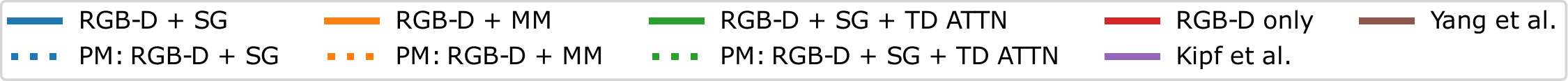}
    \caption{Quantitative and qualitative results for the \experimentthree\ task. The left two plots depict Success Rate (SR) versus environment steps over training, without (left) and with (middle) pre-mapping, respectively. We observed that \textbf{\sgnoattn} and \textbf{RGB-D + MM} slightly underperform \textbf{RGB-D only}. Adding the task-driven attention mechanism (\textbf{\sgattn}), however, results in a significant performance jump, doubling the success rate. The middle plot indicates that pre-mapping with furniture information helps guide more efficient exploration, and the scene graph models (\textbf{\sgnoattnpm} and \textbf{\sgattnpm}) are better at extracting prior information than their metric map counterparts (\textbf{PM: RGB-D + MM}). Finally, the rightmost column showcases two example agent trajectories from the \textbf{\sgattn} model: the agent is able to leverage relational information stored in the scene graph to explore efficiently (right) and backtrack from its past mistakes after entering the wrong room (left).} 
    \label{fig:exp3}
     \vspace{-2em}
\end{figure*}

From the leftmost plot of Fig.~\ref{fig:exp3} and Table~\ref{tab:exp3}, we observe that all models, including the \textbf{RGB-D only} model and the baseline models, are able to achieve some level of success. Naively introducing metric maps or scene graphs as external memory, i.e., \textbf{RGB-D + MM} and \textbf{\sgnoattn} does not boost search performance by much. We hypothesize that simply accumulating information from the past RGB-D images introduces too much noise into the training process - after all, only a very small subset of the nodes are task-relevant for the given goal description. 

\begin{table}[t]
\caption{\label{tab:exp3}\Experimentthree: Success Rate\\
and Success weighted by Path Length}
%\resizebox{0.48\textwidth}{!}{%
\resizebox{0.99\columnwidth}{!}{
\begin{tabular}{llll}
    \toprule
    {Model} & {SR}$\uparrow$  & {SPL}$\uparrow$  \\
    \midrule
    \baselineone & 0.423$\pm$0.090 & 0.221$\pm$0.068 \\
    \baselinetwo & 0.095$\pm$0.009 & 0.0302$\pm$0.0013 \\
    \midrule
    RGB-D only & 0.586$\pm$0.021 & 0.309$\pm$0.061\\
    RGB-D + MM & 0.554$\pm$0.025 & 0.273$\pm$0.025\\
    \sgnoattn & 0.458$\pm$0.135 &  0.183$\pm$0.099\\
    \sgattn & \textbf{0.879$\pm$0.048} & \textbf{0.577$\pm$0.041} \\
    \midrule
    PM: RGB-D + MM & 0.405$\pm$0.023 & 0.404$\pm$0.022\\
    \sgnoattnpm & 0.634$\pm$0.108 & 0.402$\pm$0.037\\
    \sgattnpm & \textbf{0.921$\pm$0.005} & \textbf{0.738$\pm$0.045}\\
    \bottomrule
\end{tabular}
}
\vspace{-2.5em}
\end{table}

The \textbf{\baselineone} baseline performs comparably to \textbf{\sgnoattn}, possibly because without attention, any processing of scene graphs does not yield significant improvements. Similarly, we believe that the \textbf{\baselinetwo} baseline is not able to learn at all due to its high dimensional node representation, which only serves to make learning harder.

As shown by the green curve in Fig.~\ref{fig:exp3}, the task-driven attention mechanism has a dramatic impact on the efficiency via which our method learns to extract task-relevant features from the graph. Our best model is able to converge to an 88\% success rate with only 1 million environment steps, significantly outperforming baselines and ablations.

We also analyze the effectiveness of pre-mapping the scene. Comparing the dotted lines and the solid lines in the middle plot of Fig.~\ref{fig:exp3}, we can see that pre-mapping the scene and providing the agent with furniture information leads to a sizable performance boost across all three variants of our models. The improvement over the incrementally constructed scene graph indicates that our models \textbf{\sgattn} and \textbf{\sgnoattn} can leverage prior information injected into the initial graph effectively. They also outperform the metric map counterpart (\textbf{RGB-D + MM}) with pre-mapping.

In the rightmost column of Fig.~\ref{fig:exp3}, we visualize two example agent trajectories of the \experimentthree\ task using the \textbf{\sgattn} model, where the star represents the location of the goal object. Thanks to the room-object and object-object relational information stored in the scene graph, our model can efficiently explore the scene to reach the goal object with the near-shortest path, or backtrack from past mistakes after entering the wrong room.

\section{Discussion and Limitations}
\label{s:discussion}
In this work, we show the benefits of the scene graph as a representation for Hierarchical Relational Object Navigation (HRON) tasks that require reasoning about object relations. In large, populated scenes, having a task-driven attention mechanism is essential in aggregating task-relevant information and achieving a high success rate.  

% One advantage of explicit memory models such as a scene graph over a latent representation such as weights in LSTMs~\cite{hochreiter1997long} is the ability to store and retrieve prior scene knowledge such as the room configuration, which is invariant to the task instance. Providing scene priors by populating the graph with the object information allows the scene graph based model to dramatically outperform the RGB-D only baseline across all metrics. 

%As the agent is provided the target room and associated object in the goal description, the graph model must only infer the most probable candidate locations for the missing goal object which is not provided in the pre-populated scene graph.

%In the choice task, which requires distinguishing between identical object pairs based solely on the kinematic relation, the RGBD based policy alone is not able to solve the task during the allotted training time. Building in the relational graph inductive bias allows the scene graph based model to rapidly learn to solve the task by directly leveraging the edge features which indicate which object satisfies the goal condition.

%For the relational search task, while the standard heterogeneous graph transformer network is able to perform comparably to the metric map baseline, the principal improvement of our model is provided by the attention mechanism. When compared to the ablated scene graph model that uses a global mean pooling, it is clear that the inductive bias afforded by the attention mechanism allows our model to more rapidly learn useful features for the navigation.

One advantage of explicit memory models such as scene graphs over latent representation such as weights in LSTMs~\cite{hochreiter1997long} is the ability to store and retrieve prior scene knowledge such as the room configuration and furniture placement, which is invariant to the task goal. Providing scene priors by populating the scene graph with furniture information allows the scene graph-based model with pre-mapping (\textbf{PM: \sgattn}) to dramatically outperform other baselines. %As the agent is provided the target room and associated object in the goal description, the graph model must only infer the most probable candidate locations for the missing goal object which is not provided in the pre-populated scene graph.

The scene graph representation also has several desirable properties over the metric map representation in the context of relational object search. The graph complexity scales linearly with the number of objects in the scene, rather than the size of the physical space. Given a fixed memory, a metric map must make a tradeoff between representing small objects at high resolution, or large spaces at low resolution. Furthermore, due to the 2D top-down projection, the metric map has limitations when representing occlusion or containment.

The main challenge, addressed by task-driven attention, is transforming the object-centric scene graph into the scene graph embedding provided to the policy network. Naive global pooling, as seen in the results, mixes in information from non-relevant nodes. Our task-driven attention enriches this embedding with information from task-relevant nodes.

Our method, however, is not without limitations. The scene graphs in our model are constructed using privileged information from the simulator (a perfect 3D object detector), instead of from raw visual input, which has been accomplished in previous work~\cite{wu2021scenegraphfusion}. Moreover, our problem setup is within the domain of embodied navigation without manipulation. Although the robot might collide with objects in the scene during navigation, the objects (and hence the scene graphs) stay largely static. How to leverage scene graphs and GNNs to solve mobile manipulation problems remains an active research area that is beyond the scope of this work.

%%%%%%%%%%%%%%%%%%%%%%%%%%%%%%%%%%%%%%%%%%%%%%%%%%%%%%%%%%%%%%%%%%%%%%%%%%%%%%%%



\section*{ACKNOWLEDGMENT}

This work is in part supported by ONR MURI N00014-22-1-2740, ONR MURI N00014-21-1-2801, NSF \#2120095, AFOSR YIP FA9550-23-1-0127, Stanford Institute for Human-Centered AI (HAI), Amazon, Analog Devices, Bosch, JPMC, Meta, and Salesforce.

% The preferred spelling of the word ÒacknowledgmentÓ in America is without an ÒeÓ after the ÒgÓ. Avoid the stilted expression, ÒOne of us (R. B. G.) thanks . . .Ó  Instead, try ÒR. B. G. thanksÓ. Put sponsor acknowledgments in the unnumbered footnote on the first page.

%%%%%%%%%%%%%%%%%%%%%%%%%%%%%%%%%%%%%%%%%%%%%%%%%%%%%%%%%%%%%%%%%%%%%%%%%%%%%%%%

\bibliography{ref}
\bibliographystyle{IEEEtran}

\addtolength{\textheight}{-12cm}   % This command serves to balance the column lengths

%%%%%%%%%%%%%%%%%%%%%%%%%%%%%%%%%%%%%%%%%%%%%%%%%%%%%%%%%%%%%%%%%%%%%%%%%%%%%%%%

\end{document}